\documentclass[letterpaper, 10 pt, conference]{ieeeconf}  

\IEEEoverridecommandlockouts                              
\usepackage[utf8]{inputenc}
\usepackage[T1]{fontenc}
\usepackage{amsmath}
\usepackage{graphicx}
\usepackage{cite}
\usepackage{url}

\title{\LARGE \bf
VFH+ based shared control for remotely operated mobile robots
}

\author{Pantelis Pappas$^{1}$, Manolis Chiou$^{1}$, Georgios-Theofanis Epsimos$^{1}$, Grigoris Nikolaou$^{2}$ and Rustam Stolkin$^{1}$

\thanks{$^{1}$Extreme Robotics Lab (ERL) and National Center for Nuclear Robotics (NCNR), University of Birmingham, UK}
\thanks       {\tt\small m.chiou@bham.ac.uk}%

\thanks{$^{2}$University of West Attica, Greece,
        {\tt\small nikolaou@uniwa.gr}}%
}

\begin{document}

\maketitle
\thispagestyle{empty}
\pagestyle{empty}

\begin{abstract}

This paper addresses the problem of safe and efficient navigation in remotely controlled robots operating in hazardous and unstructured environments; or conducting other remote robotic tasks. A shared control method is presented which blends the commands from a VFH+ obstacle avoidance navigation module with the teleoperation commands provided by an operator via a joypad. The presented approach offers several advantages such as flexibility allowing for a straightforward adaptation of the controller's behaviour and easy integration with variable autonomy systems; as well as the ability to cope with dynamic environments. The advantages of the presented controller are demonstrated by an experimental evaluation in a disaster response scenario. More specifically, presented evidence show a clear performance increase in terms of safety and task completion time compared to a pure teleoperation approach, as well as an ability to cope with previously unobserved obstacles.

\end{abstract}

\section{Introduction}

Research advances allowed robots to be increasingly used in time- and safety-critical applications such as robot-assisted search and rescue (SAR), hazardous environment inspection, and disaster response. Such complex and demanding applications require flexible, efficient, and robust robotic platforms. The field of remotely controlled mobile robots has been extensively researched from the point of view of traditional pure teleoperation approaches for such applications. However, recent developments in robotic technologies in both software (e.g. algorithms) and hardware (e.g. sensors an processing power) can increasingly cope with uncertainty and unstructured environments. These advances can be proved crucial in numerous disaster response and remote inspection applications such as SAR, reconnaissance in nuclear disaster sites, or any similar hazardous environments. Numerous field studies \cite{Murphy2005, Baker2004, Norton2017} have shown the lack of autonomous capabilities to be some of the major bottlenecks during robotic operations in hazardous environments. For example, the lack of autonomous capabilities in the robotic systems can lead to a drastic increase in cognitive fatigue for the human operators, and in task execution speed \cite{Casper2003}. Additionally, pure teleoperation can increase the collisions within the hazardous environment, e.g in nuclear disasters \cite{fuku_blog}. Both researchers and professional robot operators along with rescuers agree that robots that possess more autonomous capabilities could be proved beneficial during task execution. 

Variable autonomy approaches such as shared control, Human-Initiative, and Mixed-Initiative control can be adopted to tackle the aforementioned problems by combining the complementing capabilities of the human operator and the robot's AI, whilst counteracting the weaknesses of each. This is particularly important as the prevailing paradigm for robot deployment is to always have a human in-the-loop.

Reduced perception and exponentially increasing cognitive workload is a mixture that leads to a drastically degraded performance during operations. The use of shared control can allow the operator to simultaneously stay focused on many challenging tasks whilst the robot actively assists in safe navigation through the unstructured environment. For example, avoiding hazardous areas and obstacles with the utilization of shared control could be proved beneficial for robust performance during operations in difficult circumstances such as telecommunications difficulties between the robot and the operator (e.g. delay in the operator's commands). Additionally, the utilization of such a controller can contribute towards alleviating some of the control burdens and hence alleviating the cognitive workload of the operators. Post-hoc analysis of the Fukushima's nuclear disaster incident \cite{Nagatani2013} and the well documented personal experience of one of the Fukushima's robot operators as famously reported in \cite{fuku_blog}, confirm that due to high radiation levels, operations were much more difficult and the cognitive workload remained at very high levels. A characteristic example is that operators had to wear hazmat suits and 2-3 layers of gloves, greatly impairing their situation awareness and ability to control the robots. 

In this work, a shared control method is presented for safe navigation in hazardous and dynamic environments. The proposed shared controller blends the commands of a VFH+ obstacle avoidance navigation module with the motion commands provided by an operator via a joypad. The output motion commands provided by the controller result in safe trajectories capable of avoiding obstacles and hazardous areas. This work contributes by: a) showing how the conventional VFH+ autonomous navigation method can be modified to be used in the context of shared control for remote inspection and exploration tasks (i.e. modified to be goal agnostic, effective in cluttered environments, integrated with a human operator); b) showing how the modified VFH+ can be combined with the intentions of a human operator, via a conventional teleoperation interface, to achieve a shared control system; c) clearly demonstrating that the proposed shared control method outperforms conventional teleoperation in a simulated disaster response task. An advantage of our approach is its flexibility allowing for a straightforward adaptation of the resulting robot behaviour without changing the controller's core architecture. This flexibility allows for the shared controller to be easily integrated as an additional Level of Autonomy (LOA) into variable autonomy systems such as Human-Initiate (i.e. the human has the authority to initiate switches between different LOAs) \cite{Chiou2016} or Mixed-Initiative (i.e. both the human and the robot can initiate LOA switches) \cite{Chiou2020_arXiv} systems. An additional advantage is the ability to cope with dynamic environments due to the reactive nature of our shared control method.

\section{Related work}

There are a plethora of local obstacle avoidance methods reflecting the various attempts to solve the problem in the literature. However, these methods were created with autonomous robot navigation in mind. This means that shared control applications have the benefit to mend for possible deficiencies in some of these methods, while utilizing their strengths. For example, there are common problems such as "local minima traps" and "goal is unreachable" in some of the popular autonomous navigational methods. While these methods are getting constantly better they require evermore sophisticated solutions and computing power for problems that an experienced operator could easily tend to. This is one of the advantages of shared control for navigation as opposed to autonomous navigation alone. In this section, we will outline some of those methods related to our context (i.e. obstacle avoidance and shared control) and examine them from this perspective. 
 
 \subsection{Artificial Potential Fields methods}
 
 Artificial Potential Fields (PFMs or APFs) is a popular set of methods used to deal with obstacle avoidance in mobile robots and robotic manipulators, originally introduced by Khatib in 1985 \cite{1087247}. In APFs a sum of attractive and repulsive forces steers the robot appropriately and towards the target while avoiding obstacles.
 
 Recent literature is mostly focused on improving the original APF algorithm by modifying the attraction and repulsion functions \cite{Rostami2019, Sfeir2011, Gu2019}. To the best of the author's knowledge and despite their advances, the deficiencies of APFs continue to pose a problem, e.g. the navigation target being unreachable at times, the robot being susceptible to local minima traps (e.g. U-turns), and most importantly unwanted oscillatory behavior \cite{KorenY.Dept.ofMech.Eng.&Appl.Mech.MichiganUniv.AnnArborMI}.
  Additionally, two core assumptions that APFs make is that there is always a given navigational target and that the environment is always known. In contrast, in the targeted domain (e.g. disaster response), the robot often needs to build a map incrementally, without a predefined navigational goal, and while avoiding previously unobserved obstacles. Lastly, avoiding obstacles by calculating only the sum of the applied forces does not provide the required flexibility to define different behaviours for the robot to be used in shared control.
 

\subsection{Histogramic methods}
 
In an attempt to mend for the weaknesses of APFs, the Virtual Force Field (VFF) method was proposed by \textit{Borenstein and Koren}\cite{Borenstein1990}. In VFF \cite{Borenstein1990_2} the resulting direction of the robot movement is not given by the simple sum of the attractive and repulsive forces but it is determined as one of the passages in a two-dimensional Cartesian grid, called the histogram grid C. However, VFF's many shortcomings (e.g intense oscillatory behaviour, real-time mapping inefficiency) led to the development of Vector Field Histogram (VFH) \cite{Borenstein1991}. As a result, VFH robustness allows the continuous and fast motion of the mobile robot without stopping for obstacles. In VFH the obstacles do not exude a repulsive force as they do in VFF, but a vector field is built that avoids the obstacles occupying the histogram grid.

The \textbf{VFH+} algorithm  \cite{Ulrich1998} has tackled many issues of its predecessor (i.e. VFH) by taking into account the robot’s physical characteristics and kinematics. It reduces the amount of computational data needed for obstacle representation and improves on the commitment of steering decisions by the cost function. A strength of VFH+ is the fact that the robot does not oscillate when it finds more than one clear paths. As a purely reactive navigational method, VFH+ might lead the robot to dead ends. However, unlike VFH, configuring its parameters is easier and a bad configuration will not lead to catastrophic results \cite{An2004}.

In order to deal with the local nature of VFH+, Ulrich et al. \cite{Ulrich2000} proposed the \textbf{VFH*} method. The VFH* algorithm combines VFH+ and the global planner A* in order to prevent the robot from being trapped or making undesirable decisions. However, it requires careful parameter tuning and it is burdened by a heavy computational load. Additionally, it is assumed that the global planner has access to a map which is not always the case for a disaster response robot.

Babinec et al.\cite{Babinec2014} proposed the \textbf{VFH*TDT} method which is a set of modifications on the VFH+ and VFH* methods. VFH+ modifications are concerned with performance improvements in the sense of smooth movements as the reaction to obstacles. VFH* modifications are concerned to enable a simultaneous evasion of static and moving obstacles.
    
The Vector Polar Histogram (\textbf{VPH}) method \cite{An2004} leverages the accuracy of the laser range finder and reduces the number of needed steps to reliably detect the distribution of obstacles by the creation of a certainty grid for obstacle representation. \textbf{VPH+} \cite{Jianwei2007} is an extension of VPH and improves the ability to navigate in crowded environments. The algorithm groups isolated obstacle points into obstacle blocks. By classifying them as concave or non-concave the robot avoids obstacles in advance, resulting in a smoother trajectory.

Compared to the other histogramic methods, VFH+ is capable of robustly dealing with uncertainty in sensor readings. The applications we are interested in can leverage theses advantages as they can have many unforeseen circumstances that can affect the robot's sensory input in various ways, e.g. noise in laser readings due to dust in the environment. In addition, methods that implement some kind of global planning (e.g VFH*, VFH*TDT) are useful for fully autonomous robotic applications but the use of a global map and their complexity out-weight the merits of been used in a shared control context in disaster response (e.g. global map, if available, might be subject to sudden change). Lastly, some disadvantages of VFH+ (e.g. getting trapped in local minima) compared to other methods can be overcome by the shared control while avoiding more complex methods (e.g.  VFH*TDT).

\subsection{Shared control for mobile robots}

Shared control is a term regularly used for either depicting systems in which the human cooperates at some level with the robot, or to explicitly portray systems in which some type of input blending or mixing between the robot's or human's commands take place. In this paper, we will utilize this term to refer explicitly to the latter. Usually, shared control systems are concerned with the safety of the robot (i.e. avoiding collisions) and/or with minimizing the teleoperation effort of the operator/user. Although shared control is a popular approach in various robotic applications (e.g. in manipulation and grasping \cite{Adjigble2019, Javdani2018}) here we will focus on mobile robots. 

One form of shared control is safeguard operation in which the robot intervenes to stop the operator's unsafe commands in order to prevent collisions. Krotkov et al. \cite{Krotkov1996} implemented a safeguard controller to a lunar rover in order to account for time delays between commands. In the work of Fong et al \cite{Fong2001}, a safeguard controller is proposed for mobile robots deployed in unstructured environments.

Another popular application can be found in the field of robotic wheelchairs where safety, comfort, and the ability to assist users with disabilities are all strict requirements \cite{Erdogan2017}. An example that does not use histogramic methods is the work of Carlson and Demiris \cite{carlson2010increasing}. It combines safe trajectories from an AI planner with user intention prediction based on joypad commands. A shared control application for wheelchairs can be found in the work of Urdiales \textit{et al }\cite{Urdiales2007} that uses conventional APFs. They test their method in structured domestic environments. Similarly, in the work of Storms \textit{et al} \cite{Storms2017} a new obstacle representation and avoidance method based on model predictive control is presented. This approach requires human operators' models for shared control to function. Other shared control approaches include haptic feedback and personalized assistance by human demonstration that have proven to increase performance and joystick jerkiness \cite{Kucukyilmaz2018} \cite{Saeidi2017}.

Most related to our paper is the work of Bell \textit{et al} \cite{Bell1994} in which the VFH method overrides unsafe operator commands. However, this approach fails to navigate in narrow passages (e.g a doorway), making the use of a separate control mode for such cases a necessity. Another related shared control implementation can be seen in the work of Chen \textit{et al} \cite{Chen2019}. They implement VFH+ assisted gesture and voice control for an omnidirectional mobile robot for structured domestic environments. 

Many of the above approaches are concerned with domestic scenarios in structured environments and in many cases, the operator is physically situated in the same space as the robot (e.g. in robotic wheelchairs). Contrary, our work is focused on remotely controlled robots in applications that involve unstructured and dynamic environments such as SAR and a variety of performance degrading conditions (e.g. communication delays). Additionally, some of the above methods prevent unsafe user commands in an intrusive way (e.g safeguard modes). In contrast, we leverage the capability of the VFH+ method and shared control for smooth command blending instead of overriding or interrupting the operator's commands.

\section{VFH+ based shared control}

The problem addressed here is the design of a shared controller for safe navigation (i.e. avoiding obstacles and hazardous areas) of remotely operated mobile robots. The robot's autonomous obstacle avoidance commands $U_r$ and the operator's command $U_h$ acts as inputs to the controller which blends them, and outputs safe motion commands $U_f$. These output commands $U_f$ are fed to the robot actuators and result in a safe trajectory. The robot's input $U_r$ is the velocity commands produced by a VFH+ obstacle avoidance module and expresses velocity in free space broken into its linear and angular parts. The same applies to the human operator's input $U_h$ which is produced via joypad commands.

\subsection{VFH+ obstacle avoidance module}

Here, briefly and for completeness, our VFH+ obstacle avoidance module is presented. Modifications and differences from the original VFH+ will be mentioned where relevant.

First, a histogram grid is defined as a square-shaped active window ($w_s  \times  w_s$). The grid is updated in real time using laser range finder sensors and by taking the laser's maximum measurement range $d_{max}$ we can determine the grid's dimensions based on (\ref{eq: d_max}):

\begin{equation}
  d_{max} = \frac{\sqrt{2}( w_s - 1)}{2} 
  \label{eq: d_max}
\end{equation}

In our system the window we use is a $60 x 60$ cell histogram grid (i.e. $w_s=60)$) with a cell size of 100mm that gives us an active square window $C_{\alpha}$ of $4m^2$ (i.e. $4m$ in each direction from the robot's center).

Then, \textit{the primary polar histogram} is derived from the data collected by the laser and by calculating each cell in the active window into vectors with the attributes of magnitude and direction.

As a standard procedure in VFH+, the obstacles are enlarged by a radius $r_{r+s} = r_r + d_s$ where, $d_s$ is the minimum distance between the robot and an obstacle and $r_r$ the robot radius.
We heuristically defined the robot radius for the purpose of maneuvering more fluently in narrow corridors, compared to the standard VFH+. A factor to this decision was that the Husky UGV that was used for the experiments is a relatively long rectangular shaped vehicle. This results in a radius of roughly $530mm$ which is considerably large for the robot. It was heuristically found that a $26\%$ decrease in radius (i.e. a radius of $400mm$), yields much better results. Since there was a reduction in the robot radius, we empirically increase the $r_r$ value by $10\%$ in the $r_{r+s}$ operation for added safety, regardless of our chosen safety distance $d_s$.

The phases of \textit{binary polar histogram} where the obstacle representation is validated from misreadings and the \textit{masked polar histogram} where candidate directions are generated are calculated as the original method entails.

Lastly, the VFH+ cost function looks at all openings in the masked polar histogram and calculates candidate directions for the robot to move. The candidate direction $k_d$ with the lowest cost is then chosen to be the new direction of motion $\phi_d = k_d \cdot \alpha $ , where $\alpha$ is the angular resolution. It distinguishes between \textit{wide} and \textit{narrow} openings. If the right and left sectors $(k_r,k_l)$ is smaller than $s_{max}$ sectors the opening is considered narrow. In that case there is only one candidate direction, that steers the robot through the center of the opening:

\begin{equation}
    c_d = \frac{k_r + k_l}{2}
\end{equation}

In the case of a wide opening there are two candidate directions $c_r$ to the right and $c_l$ to the left side of the opening:

\begin{align}
c_r&= k_r + \frac{s_{max}}{2}\\
c_l&= k_l - \frac{s_{max}}{2} \nonumber
\end{align}

The original VFH+ method requires a candidate direction equation reflecting the direction of the navigational goal (\ref{Equation 16}). In contrast, given the requirements of targeted domain (e.g. exploration in disaster sites) our approach offers a goal agnostic VFH+ method. This is achieved by setting a constant $90^{\circ}$ angle for its goal, the forward moving direction of the robot. This means that the robot will not make turns in the absence of obstacles by trying to reach a target. The appropriate direction is selected by the VFH+ cost function.

\begin{align} \label{Equation 16}
c_t&=k_t && \text{if $k\in[c_r,c_l]$}
\end{align}

\subsection{Shared control}

\begin{figure}
  \centering
  \includegraphics[width=0.7\columnwidth]{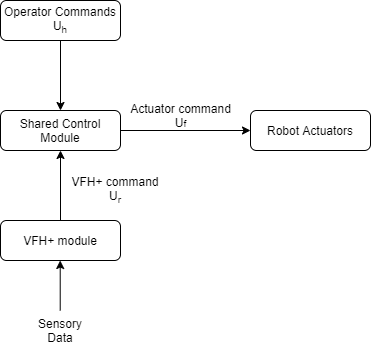}
  \caption{The block diagram of the shared controller.}
  \label{fig:shared_contol_diagram}
    
\end{figure}

The shared controller is responsible for blending the VFH+ module's velocity commands $U_r$ and the operator's joypad commands $U_h$, resulting in safe $U_f$ output velocities for the robot to follow (see Fig. \ref{fig:shared_contol_diagram}).

Mathematically, shared control is often seen as arbitration of two policies (i.e. policy blending) \cite{Dragan2013}. In our case the arbitration function that expresses this blending is the following: 

\begin{equation}
  \label{merging_exp}
  U_{f} = \alpha(\text{·}) U_{h} + ( 1 - \alpha(\text{·}) ) U_{r}
\end{equation}

The influence of $U_{h}$ and $U_{r}$ is adjusted by the arbitrator function $\alpha$(·). The function's values can range from 0.0 to 1.0 and can be adapted on the application, conditions and controller's behavior requirements. For example, if an operator requires more assistance due to high workload then with the use of the appropriate adaptive function the level of $U_{r}$ can be modified (e.g. become the more dominant input). In our case, $\alpha$(·) remains a constant for the purposes of this paper expressing the linear blending between $U_{h}$ and $U_{r}$. The value $\alpha(\text{·}) = 0.5$ was chosen heuristically and it practically means that both robot's and operator's commands are contributing equally to the final velocity vector $U_{f}$. The blending node is responsible for producing the final vector that will be driven to the robot's actuators. The operations are as follows:
\begin{equation}
  u_{linear} = \alpha(\text{·}) u_{h_{linear}} + ( 1 - \alpha(\text{·})) u_{r_{linear}}
\end{equation}

\begin{equation}
u_{angular} = \alpha(\text{·}) u_{h_{angular}} +( 1 - \alpha(\text{·}) ) u_{r_{angular}}
\end{equation} 

\begin{equation}
    u_{f} = u_{linear} + u_{angular}
\end{equation}

The shared controller proposed can run in the background without affecting the behavior of the robot. It can be activated in runtime and on-demand (e.g. with the press of a button by the operator in Human-Initiative control or by the robot in Mixed-Initiative control systems) in situations that can be beneficial.

\section{Experimental evaluation}

\begin{figure}
	\centering
	\includegraphics[width=0.99\columnwidth]{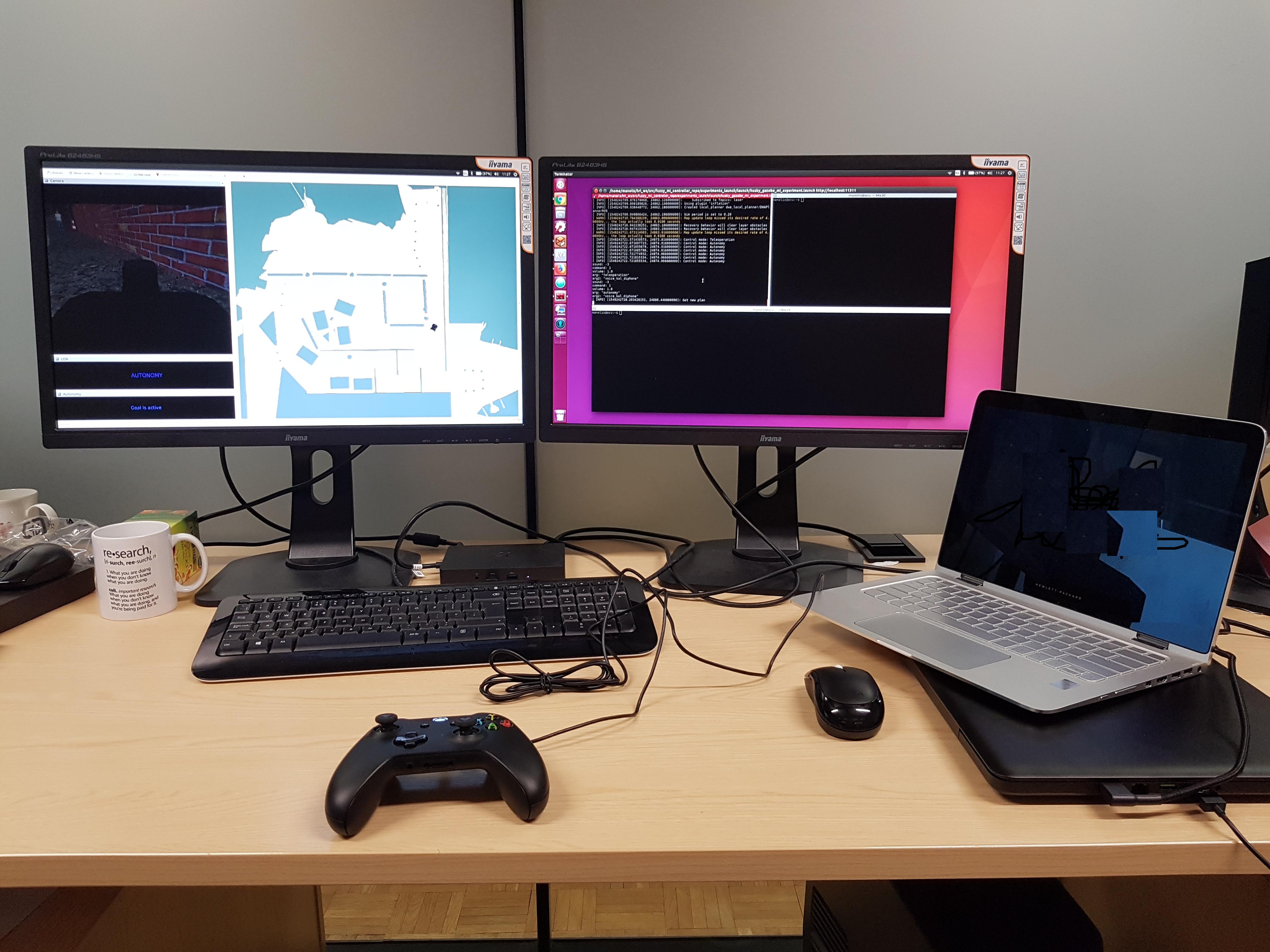}
	\caption{The Operator Control Unit (OCU) composed of a laptop, a joypad, and a screen showing the GUI.} 
	\label{fig:ocu}
\end{figure}

\begin{figure}
	\centering
	\includegraphics[width=0.99\columnwidth]{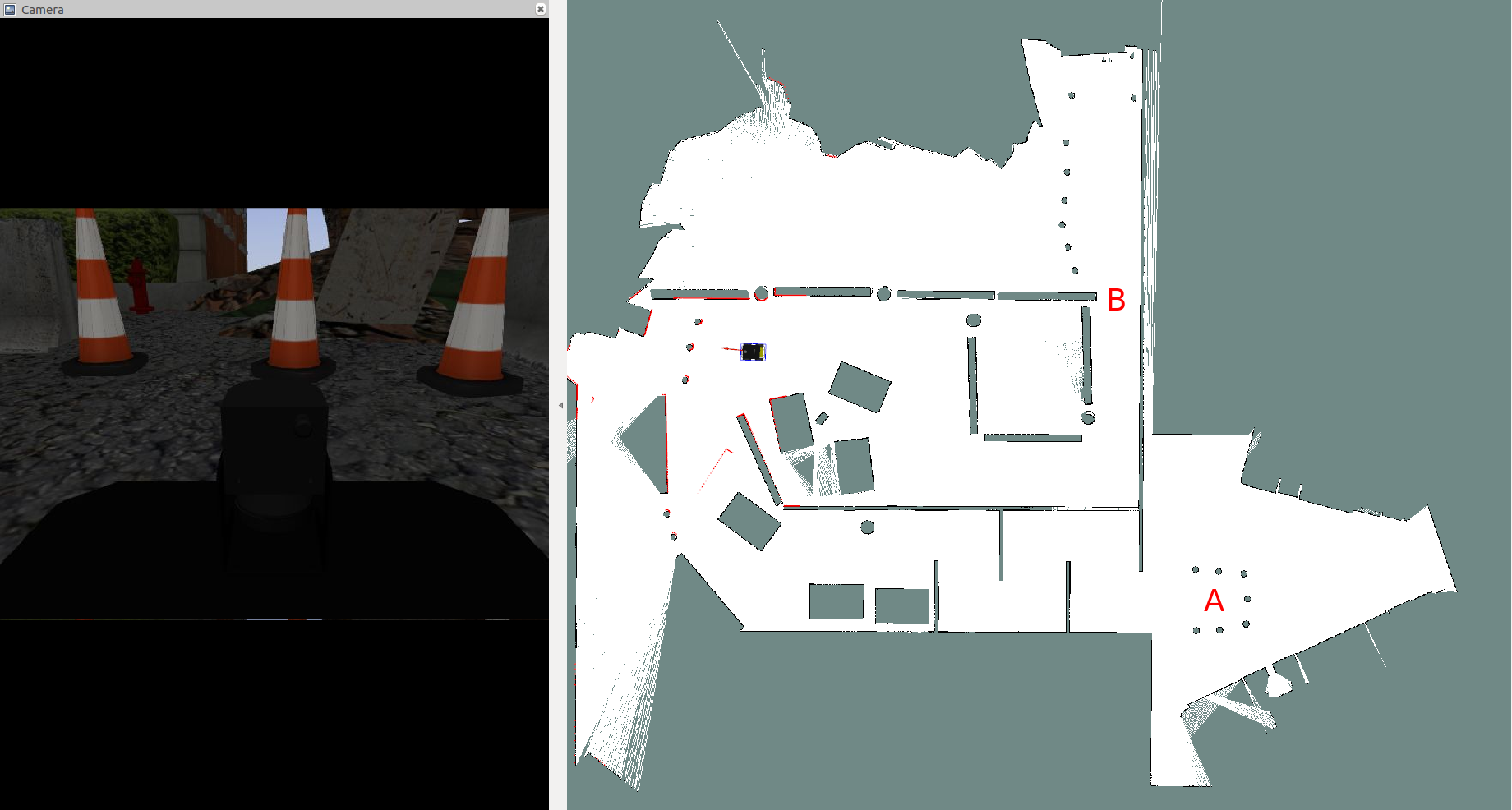}
	\caption{\textbf{Left:} video feed from the camera. \textbf{Right:} The map (as created by SLAM) showing the pose of the robot, the obstacles’ laser reflections (red), and the walls (black). In the map, the task was to navigate from point A to point B.} 
	\label{fig:gui}
\end{figure}

\begin{figure}
	\centering
	\includegraphics[width=0.99\columnwidth]{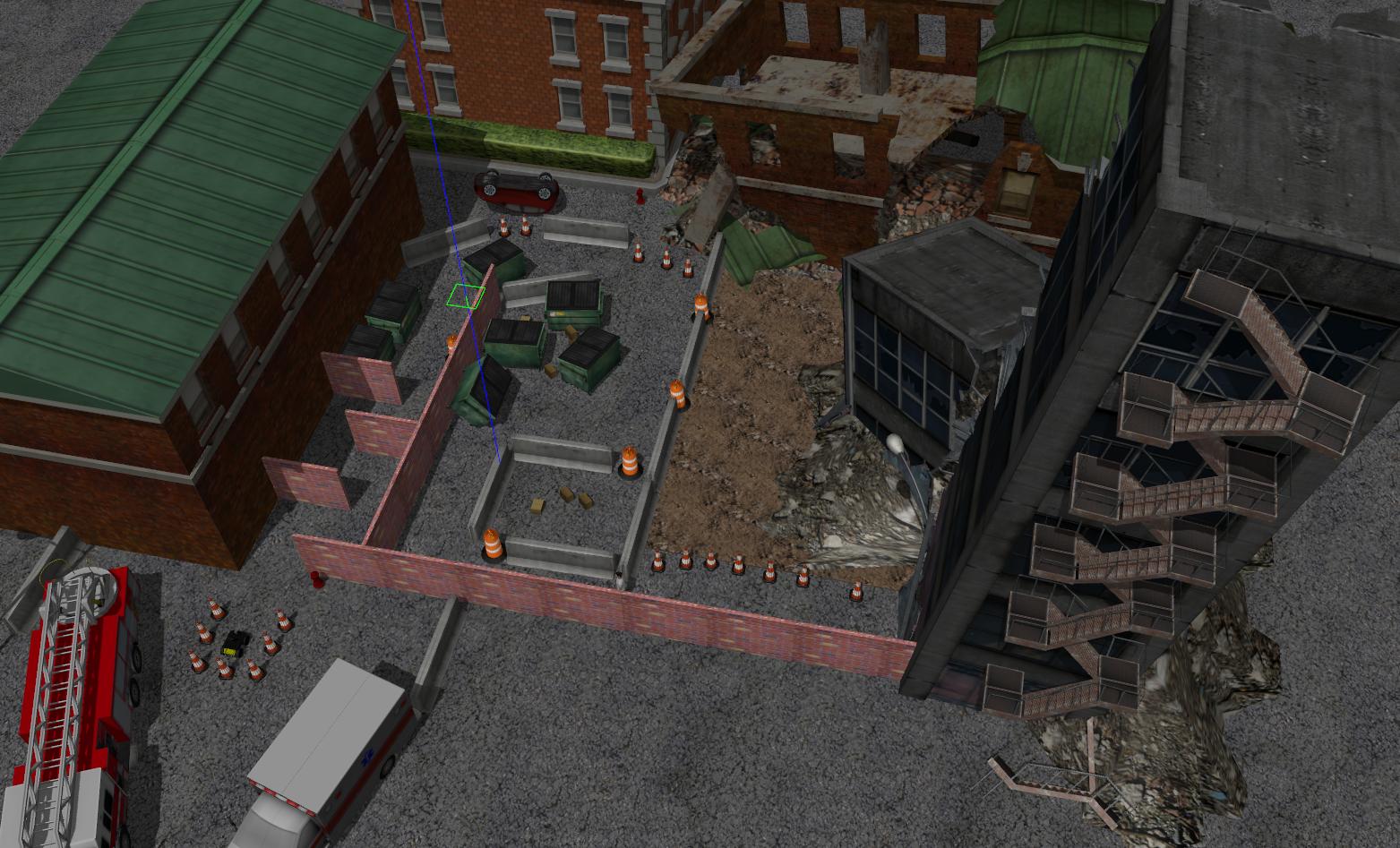}
	\caption{The test arena used, simulating a SAR scenario.} 
	\label{fig:arena}
\end{figure}

An experiment was conducted to evaluate the performance of the proposed shared control method. For the experiment, a realistically simulated SAR test arena (i.e. SAR environment) with dimensions of approximately $24m \times 24m$ (see Fig. \ref{fig:gui} and \ref{fig:arena}) was created. Gazebo, a high fidelity robotic simulator, was used to simulate the environment and the robotic system. Gazebo simulator uses an advanced physics engine and creates realistic environments and stimuli for the operators, as it can be seen in Figures \ref{fig:arena} and \ref{fig:gui}. The simulated robot was equipped with a laser range finder and an RGB camera. It was controlled via an Operator Control Unit (OCU) (see Fig. \ref{fig:ocu}). The OCU was composed of a mouse, a joypad for the operator's commands, and a laptop running the software and a screen showing the Graphical User Interface (GUI) (see Fig. \ref{fig:gui}). The software used was developed in the Robot Operating System (ROS). The repository\footnote{Extreme Robotics Lab GitHub repository: \url{https://github.com/uob-erl/shared_control}} containing the ROS code for the VFH+ based shared control described in this paper, is provided under MIT license.

The simulation was used in order to avoid the introduction of confounding factors from a real-world robot deployment and to improve the repeatability of the experiment. This is especially true given the complexity of the experiment and the size of the testing arena. For example, wireless communication with the robot failing due to unpredictable signal degradation can act as a confounding factor that negatively affects our controlled experiment. Additionally, our system is a remotely controlled robot and hence the control interface (i.e. the OCU with the joypad and the GUI) remains the same in a real-world deployment.

\subsection{Experimental protocol}

Three expert robot operators were given the task of navigating from point A to point B (see Fig. \ref{fig:gui}) as fast and as safely as possible. Each of the three operators ran 8 trials of the task in total; 4 times using pure teleoperation and 4 times using the shared controller. In order to counterbalance and to minimize the learning effect, the trials were run in an alternating pattern.

A SLAM generated map was given to the system and the operator (i.e. via the GUI) at the beginning of the trial. However, before each trial, the experimenter placed randomly several additional obstacles in the arena for which the operators were not aware and they were only visible in the map as laser reflections and only in close proximity. This was in order to introduce a dynamic environment effect common in disaster response scenarios. Care was taken for the difficulty of the overall task to remain the same between trials regardless of the positions of the random obstacles. Additionally, to further the realism of the task, a $1 sec$ delay was introduced to the operator's commands. Also, the camera's image transmission bandwidth was reduced to 2.5Hz. The delayed commands and video feedback are common communication difficulties during remote operations e.g. in \cite{Lane2001,Gomez-de-Gabriel2000} and in DARPA DRC Finals \cite{Norton2017}. 

\subsection{Results}
Two performance metrics were measured: the time-to-completion reflecting how much time it took to complete the task; and the number of collisions with the environment (i.e. obstacles). The data were tested for normality with the Shapiro-Wilk test and in addition, were inspected visually. Time-to-completion data conformed to a normal distribution and hence a paired sample t-test was used to compare the means. The number of collisions data were not normally distributed and hence the Wilcoxon signed-rank test was used. We consider a result to be statistically significant when it yields a $p$ value less than 0.05.

\begin{figure} 
  \centering
  \includegraphics[width=\linewidth]{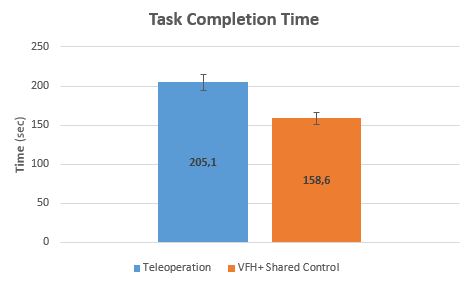}
  \caption{Task completion time bar chart. The error bars indicate the standard error.}
  \label{fig:time_barchart}
\end{figure}

\textbf{Task completion time (secs):} Operators in shared control completed the task significantly ($t(11) = 10.209, p < .001.$) quicker ($M = 158.3, SD = 10.7$) compared to teleoperation ($M = 205.1, SD = 19.9$), see Fig. \ref{fig:time_barchart}.

\begin{figure} 
  \centering
  \includegraphics[width=\linewidth]{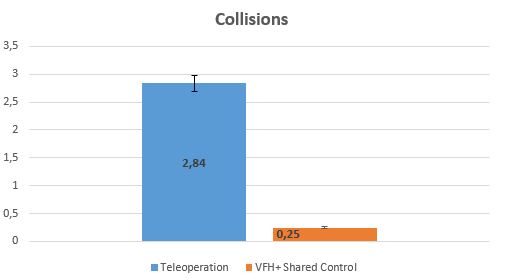}
  \caption{Number of collisions bar chart.The error bars indicate the standard error.}
  \label{fig:collisions_barchart}
 \end{figure}

\textbf{Number of collisions:} Operators in shared control had significantly ($z = -2.82, p < .01.$) fewer collisions ($M = 0.25, SD = 0.45$) compared to teleoperation ($M = 2.8, SD = 2.29$), see Fig. \ref{fig:collisions_barchart}.

\subsection{Discussion}

Our experimental evaluation has shown that shared control improves performance as the navigation task is completed by \textbf{30\%} faster and there are \textbf{120\%} fewer collisions compared to teleoperation. The nature of our experiment is meant to reflect a realistic scenario where operators are under pressure in degrading robot control conditions. While the subjects were all expert robot operators (i.e. extensive experience in operating similar robots), the task at hand was significantly hard. A factor that must be considered is that the human agent is remotely situated and due to the increased workload, his Situational Awareness is greatly reduced \cite{riley2006effects}. All the aforementioned factors were present during our experiment and results emphasize the need and the advantages of the proposed shared control system.

In Teleoperation mode, a significant effort had to be put by each subject to navigate the robot in the arena, as the lag in the video feedback and the control commands can be mentally draining. One of the factors that contributed to the high task completion time and a high number of collisions in teleoperation was the fact that users failed to make precise movements due to the input and visual latency. To deal with this, subjects had to adopt a stop-and-wait strategy and avoid complicated commands. However, they frequently failed to anticipate the effect of their commands and often seem to overcompensate with additional commands in order to correct their course. This often resulted in self-induced oscillatory behaviour and collisions, even in relatively simple turns or in straight corridors. Another factor that might explain the poor performance of teleoperation, is the gradual exhaustion of operators during trials as a result of the above mention difficulties.

Shared control's improved performance can be attributed to the reduction of necessary commands in order to navigate, as well as the reduction of their complexity. Despite the commands input latency, in practice, the subjects only needed to input the general direction they wished to execute. The module's self-correction absolves the operator from complex maneuvers and the robot would follow the instructed direction of its course. Additionally, the reactive nature of our shared control method is able to cope with dynamically changing environments. The randomly placed obstacles did not seem to degrade performance regardless of their placement. Anecdotal evidence suggests that the robot's actions were not contrary to the operator's commands. The operators largely felt like the robot was driving and steering itself, which yields better results compared to teleoperation, despite the subject's exhaustion. 

In this work, the arbitration function $\alpha$(·) was a constant denoting an equal amount of commands blending from the operator and the robot. However, there are situations that can benefit from lower values of $\alpha$(·) (i.e. the robot commands are the dominant input). For example the operator having a high workload or wireless communication issues. On the other hand, in situation that the human input might be more beneficial, such as driving through tight/narrow corridors or other precise maneuvering (e.g. the robot being stuck), higher values of $\alpha$(·) would be useful. An arbitration function that adapts the level of robot's assistance to the circumstances will contribute the most towards performance and should be the subject of future research. Our anecdotal observations suggest that a meaningful range for $\alpha$(·) is between $0.3$ and $0.7$.


The results presented here further contribute to related literature that has shown that delayed control inputs can drastically increase task completion time from $160\%$ up to $480\%$ \cite{Lane2001} in mobile robots, depending on the severity of the latency. Additionally, we contribute further evidence of the advantages of shared control in outperforming conventional teleoperation under communication delays in accordance with similar findings from Storms et al. \cite{Storms2017}.

Lastly, based on the known shortcomings of VFH+ (e.g. local minima), the occasional unsatisfactory performance was expected. However, by using the proposed shared control method in practice, the robot did not show signs of getting trapped in local minima or signs of struggle in narrow openings/corridors. This is possibly because the input from the operator directly affects the forwards and backward movements of the robot, and in the context of shared control compensated for the shortcomings of VFH+ by overruling situations where the robot would consider or lead itself to being trapped. This is a scenario that demonstrates the complementing capabilities of human operators and robots (i.e. autonomous capabilities) with the two agents complementing each other's strengths and highlighting the importance of collaborating as a human-robot team in demanding tasks.

\section{Conclusions}

This paper presented a VFH+ based shared control method for remotely controlled mobile robots. The method was evaluated in a navigation task in a simulated disaster site. The difficulty of controlling such robots with conventional teleoperation in performance degraded conditions such as communication delays is demonstrated by the resulting poor performance in the task. In contrast, the proposed shared control method provided a significant improvement in terms of safety and task completion time compared to teleoperation. 

In future work, we aim to further develop the shared control method by incorporating a dynamic arbitration function. This function can dynamically arbitrate how much control should be given to the robot or the operator. This can be based on the explicit use of operator intent or operator state (e.g. operator having a high workload). Additionally, further testing with a variety of performance degrading conditions both for the operation and the robot (e.g external distractions or additional sensor noise) should be made in order to determine potential weaknesses that we have yet to come across. The next milestone of our work is towards a shared control method that is able to avoid hazardous regions entirely, e.g. a region with high radiation levels that can destroy the robot's electronics or areas of extreme heat. In essence, the integration of our method with a variety of sensors (e.g. radiation sensors or infrared sensors) and map representations (e.g. radiation map).

Lastly, our shared control module was implemented as a Level of Autonomy (LOA) to be used in variable autonomy systems that switch on-demand between different LOAs. Hence, future work should explore the integration and merits of the proposed shared control method as an additional LOA used in Human-Initiative and Mixed-Initiative robotic systems.

\section*{Acknowledgment}
This work was supported by the following grants of UKRI-EPSRC: EP/P017487/1  (Remote  Sensing  in  Extreme Environments); EP/R02572X/1 (National  Centre  for  Nuclear  Robotics); EP/P01366X/1 (Robotics for  Nuclear Environments). Stolkin was also sponsored by a Royal Society Industry Fellowship.


\bibliographystyle{IEEEtran}
\bibliography{IEEEabrv, ./bibliography/root}

\begin{thebibliography}{10}
\providecommand{\url}[1]{#1}
\csname url@samestyle\endcsname
\providecommand{\newblock}{\relax}
\providecommand{\bibinfo}[2]{#2}
\providecommand{\BIBentrySTDinterwordspacing}{\spaceskip=0pt\relax}
\providecommand{\BIBentryALTinterwordstretchfactor}{4}
\providecommand{\BIBentryALTinterwordspacing}{\spaceskip=\fontdimen2\font plus
\BIBentryALTinterwordstretchfactor\fontdimen3\font minus
  \fontdimen4\font\relax}
\providecommand{\BIBforeignlanguage}[2]{{%
\expandafter\ifx\csname l@#1\endcsname\relax
\typeout{** WARNING: IEEEtran.bst: No hyphenation pattern has been}%
\typeout{** loaded for the language `#1'. Using the pattern for}%
\typeout{** the default language instead.}%
\else
\language=\csname l@#1\endcsname
\fi
#2}}
\providecommand{\BIBdecl}{\relax}
\BIBdecl

\bibitem{Murphy2005}
R.~R. Murphy and J.~L. Burke, ``Up from the rubble: Lessons learned about hri
  from search and rescue,'' in \emph{Proceedings of the Human Factors and
  Ergonomics Society Annual Meeting}, vol.~49, no.~3.\hskip 1em plus 0.5em
  minus 0.4em\relax SAGE Publications Sage CA: Los Angeles, CA, 2005, pp.
  437--441.

\bibitem{Baker2004}
M.~Baker, R.~Casey, B.~Keyes, and H.~A. Yanco, ``Improved interfaces for
  human-robot interaction in urban search and rescue,'' in \emph{2004 IEEE
  International Conference on Systems, Man and Cybernetics (IEEE Cat. No.
  04CH37583)}, vol.~3.\hskip 1em plus 0.5em minus 0.4em\relax IEEE, 2004, pp.
  2960--2965.

\bibitem{Norton2017}
A.~Norton, W.~Ober, L.~Baraniecki, E.~McCann, J.~Scholtz, D.~Shane, A.~Skinner,
  R.~Watson, and H.~Yanco, ``Analysis of human--robot interaction at the darpa
  robotics challenge finals,'' \emph{The International Journal of Robotics
  Research}, vol.~36, no. 5-7, pp. 483--513, 2017.

\bibitem{Casper2003}
J.~{Casper} and R.~R. {Murphy}, ``Human-robot interactions during the
  robot-assisted urban search and rescue response at the world trade center,''
  \emph{IEEE Transactions on Systems, Man, and Cybernetics, Part B
  (Cybernetics)}, vol.~33, no.~3, pp. 367--385, 2003.

\bibitem{fuku_blog}
E.~Guizzo, ``Fukushima robot operator writes tell-all blog,'' \emph{IEEE
  Spectrum}, vol.~23, 2011.

\bibitem{Nagatani2013}
K.~Nagatani, S.~Kiribayashi, Y.~Okada, K.~Otake, K.~Yoshida, S.~Tadokoro,
  T.~Nishimura, T.~Yoshida, E.~Koyanagi, M.~Fukushima \emph{et~al.},
  ``Emergency response to the nuclear accident at the fukushima daiichi nuclear
  power plants using mobile rescue robots,'' \emph{Journal of Field Robotics},
  vol.~30, no.~1, pp. 44--63, 2013.

\bibitem{Chiou2016}
M.~Chiou, R.~Stolkin, G.~Bieksaite, N.~Hawes, K.~L. Shapiro, and T.~S.
  Harrison, ``Experimental analysis of a variable autonomy framework for
  controlling a remotely operating mobile robot,'' in \emph{2016 IEEE/RSJ
  International Conference on Intelligent Robots and Systems (IROS)}.\hskip 1em
  plus 0.5em minus 0.4em\relax IEEE, 2016, pp. 3581--3588.

\bibitem{Chiou2020_arXiv}
\BIBentryALTinterwordspacing
M.~Chiou, N.~Hawes, and R.~Stolkin, ``{Mixed-Initiative variable autonomy for
  remotely operated mobile robots},'' 2020. [Online]. Available:
  \url{https://arxiv.org/abs/1911.04848}
\BIBentrySTDinterwordspacing

\bibitem{1087247}
O.~{Khatib}, ``Real-time obstacle avoidance for manipulators and mobile
  robots,'' in \emph{Proceedings. 1985 IEEE International Conference on
  Robotics and Automation}, vol.~2, 1985, pp. 500--505.

\bibitem{Rostami2019}
S.~M.~H. Rostami, A.~K. Sangaiah, J.~Wang, and X.~Liu, ``Obstacle avoidance of
  mobile robots using modified artificial potential field algorithm,''
  \emph{EURASIP Journal on Wireless Communications and Networking}, vol. 2019,
  no.~1, p.~70, 2019.

\bibitem{Sfeir2011}
J.~{Sfeir}, M.~{Saad}, and H.~{Saliah-Hassane}, ``An improved artificial
  potential field approach to real-time mobile robot path planning in an
  unknown environment,'' in \emph{2011 IEEE International Symposium on Robotic
  and Sensors Environments (ROSE)}, 2011, pp. 208--213.

\bibitem{Gu2019}
X.~{Gu}, M.~{Han}, W.~{Zhang}, G.~{Xue}, G.~{Zhang}, and Y.~{Han},
  ``Intelligent vehicle path planning based on improved artificial potential
  field algorithm,'' in \emph{2019 International Conference on High Performance
  Big Data and Intelligent Systems (HPBD IS)}, 2019, pp. 104--109.

\bibitem{KorenY.Dept.ofMech.Eng.&Appl.Mech.MichiganUniv.AnnArborMI}
Y.~Koren, J.~Borenstein \emph{et~al.}, ``Potential field methods and their
  inherent limitations for mobile robot navigation.'' in \emph{ICRA}, vol.~2,
  1991, pp. 1398--1404.

\bibitem{Borenstein1990}
J.~{Borenstein} and Y.~{Koren}, ``Real-time obstacle avoidance for fast mobile
  robots in cluttered environments,'' in \emph{Proceedings., IEEE International
  Conference on Robotics and Automation}, 1990, pp. 572--577 vol.1.

\bibitem{Borenstein1990_2}
J.~Borenstein and Y.~Koren, ``Teleautonomous guidance for mobile robots,''
  \emph{IEEE Transactions on Systems, Man, and Cybernetics}, vol.~20, no.~6,
  pp. 1437--1443, 1990.

\bibitem{Borenstein1991}
J.~Boernstein and Y.~Koren, ``The vector field histogram-fast obstacle
  avoidance for mobile robots,'' \emph{IEEE Transaction on Robotics
  Automation}, vol.~7, no.~3, pp. 278--288, 1991.

\bibitem{Ulrich1998}
I.~{Ulrich} and J.~{Borenstein}, ``Vfh+: reliable obstacle avoidance for fast
  mobile robots,'' in \emph{Proceedings. 1998 IEEE International Conference on
  Robotics and Automation (Cat. No.98CH36146)}, vol.~2, 1998, pp. 1572--1577
  vol.2.

\bibitem{An2004}
{Dong An} and {Hong Wang}, ``Vph: a new laser radar based obstacle avoidance
  method for intelligent mobile robots,'' in \emph{Fifth World Congress on
  Intelligent Control and Automation (IEEE Cat. No.04EX788)}, vol.~5, 2004, pp.
  4681--4685 Vol.5.

\bibitem{Ulrich2000}
I.~{Ulrich} and J.~{Borenstein}, ``Vfh/sup */: local obstacle avoidance with
  look-ahead verification,'' in \emph{Proceedings 2000 ICRA. Millennium
  Conference. IEEE International Conference on Robotics and Automation.
  Symposia Proceedings (Cat. No.00CH37065)}, vol.~3, 2000, pp. 2505--2511
  vol.3.

\bibitem{Babinec2014}
A.~Babinec, F.~Ducho{\v{n}}, M.~Dekan, P.~P{\'a}szt{\'o}, and M.~Kelemen,
  ``Vfh* tdt (vfh* with time dependent tree): A new laser rangefinder based
  obstacle avoidance method designed for environment with non-static
  obstacles,'' \emph{Robotics and autonomous systems}, vol.~62, no.~8, pp.
  1098--1115, 2014.

\bibitem{Jianwei2007}
J.~{Gong}, Y.~{Duan}, Y.~{Man}, and G.~{Xiong}, ``Vph+: An enhanced vector
  polar histogram method for mobile robot obstacle avoidance,'' in \emph{2007
  International Conference on Mechatronics and Automation}, 2007, pp.
  2784--2788.

\bibitem{Adjigble2019}
M.~Adjigble, N.~Marturi, V.~Ortenzi, and R.~Stolkin, ``An assisted
  telemanipulation approach: combining autonomous grasp planning with haptic
  cues.'' in \emph{IROS}, 2019, pp. 3164--3171.

\bibitem{Javdani2018}
S.~Javdani, H.~Admoni, S.~Pellegrinelli, S.~S. Srinivasa, and J.~A. Bagnell,
  ``Shared autonomy via hindsight optimization for teleoperation and teaming,''
  \emph{The International Journal of Robotics Research}, vol.~37, no.~7, pp.
  717--742, 2018.

\bibitem{Krotkov1996}
E.~Krotkov, R.~Simmons, F.~Cozman, and S.~Koenig, ``Safeguarded teleoperation
  for lunar rovers: From human factors to field trials,'' in \emph{IEEE
  Planetary Rover Technology and Systems Workshop}, vol.~26, 1996, p.~28.

\bibitem{Fong2001}
T.~Fong, C.~Thorpe, and C.~Baur, ``A safeguarded teleoperation controller,'' in
  \emph{IEEE International Conference on Advanced Robotics (ICAR)}, no. CONF,
  2001.

\bibitem{Erdogan2017}
A.~Erdogan and B.~D. Argall, ``The effect of robotic wheelchair control
  paradigm and interface on user performance, effort and preference: an
  experimental assessment,'' \emph{Robotics and Autonomous Systems}, vol.~94,
  pp. 282--297, 2017.

\bibitem{carlson2010increasing}
T.~{Carlson} and Y.~{Demiris}, ``Increasing robotic wheelchair safety with
  collaborative control: Evidence from secondary task experiments,'' in
  \emph{2010 IEEE International Conference on Robotics and Automation}, 2010,
  pp. 5582--5587.

\bibitem{Urdiales2007}
C.~Urdiales, A.~Poncela, I.~Sanchez-Tato, F.~Galluppi, M.~Olivetti, and
  F.~Sandoval, ``Efficiency based reactive shared control for collaborative
  human/robot navigation,'' in \emph{2007 IEEE/RSJ International Conference on
  Intelligent Robots and Systems}.\hskip 1em plus 0.5em minus 0.4em\relax IEEE,
  2007, pp. 3586--3591.

\bibitem{Storms2017}
J.~Storms, K.~Chen, and D.~Tilbury, ``A shared control method for obstacle
  avoidance with mobile robots and its interaction with communication delay,''
  \emph{The International Journal of Robotics Research}, vol.~36, no. 5-7, pp.
  820--839, 2017.

\bibitem{Kucukyilmaz2018}
A.~Kucukyilmaz and Y.~Demiris, ``Learning shared control by demonstration for
  personalized wheelchair assistance,'' \emph{IEEE transactions on haptics},
  vol.~11, no.~3, pp. 431--442, 2018.

\bibitem{Saeidi2017}
H.~Saeidi, J.~R. Wagner, and Y.~Wang, ``A mixed-initiative haptic teleoperation
  strategy for mobile robotic systems based on bidirectional computational
  trust analysis,'' \emph{IEEE Transactions on Robotics}, vol.~33, no.~6, pp.
  1500--1507, 2017.

\bibitem{Bell1994}
D.~A. {Bell}, J.~{Borenstein}, S.~P. {Levine}, Y.~{Koren}, and J.~{Jaros}, ``An
  assistive navigation system for wheelchairs based upon mobile robot obstacle
  avoidance,'' in \emph{Proceedings of the 1994 IEEE International Conference
  on Robotics and Automation}, 1994, pp. 2018--2022 vol.3.

\bibitem{Chen2019}
W.~Chen, C.~Yang, and Y.~Feng, ``Shared control for omnidirectional mobile
  robots,'' in \emph{2019 Chinese Control And Decision Conference
  (CCDC)}.\hskip 1em plus 0.5em minus 0.4em\relax IEEE, 2019, pp. 6185--6190.

\bibitem{Dragan2013}
A.~D. Dragan and S.~S. Srinivasa, ``A policy-blending formalism for shared
  control,'' \emph{The International Journal of Robotics Research}, vol.~32,
  no.~7, pp. 790--805, 2013.

\bibitem{Lane2001}
J.~C. Lane, C.~R. Carignan, and D.~L. Akin, ``Time delay and communication
  bandwidth limitation on telerobotic control,'' in \emph{Mobile Robots XV and
  Telemanipulator and Telepresence Technologies VII}, vol. 4195.\hskip 1em plus
  0.5em minus 0.4em\relax International Society for Optics and Photonics, 2001,
  pp. 405--419.

\bibitem{Gomez-de-Gabriel2000}
J.~G{\'o}mez-de Gabriel and A.~Ollero, ``Overcoming communication delay in
  vehicle teleoperation,'' \emph{IFAC Proceedings Volumes}, vol.~33, no.~25,
  pp. 131--136, 2000.

\bibitem{riley2006effects}
J.~M. Riley and L.~D. Strater, ``Effects of robot control mode on situation
  awareness and performance in a navigation task,'' in \emph{Proceedings of the
  Human Factors and Ergonomics Society annual meeting}, vol.~50, no.~3.\hskip
  1em plus 0.5em minus 0.4em\relax SAGE Publications Sage CA: Los Angeles, CA,
  2006, pp. 540--544.

\end{thebibliography}

\end{document}